\documentclass[runningheads]{llncs}
\usepackage[T1]{fontenc}
\usepackage{graphicx}
\usepackage{url}
\usepackage{comment}
\usepackage{amsmath}
\usepackage{amssymb}
\usepackage{amsfonts}       
\usepackage[utf8]{inputenc} 
\usepackage{booktabs}       
\usepackage{nicefrac}       
\usepackage{microtype}      
\usepackage{xcolor}         
\usepackage{epsfig}
\usepackage{soul}
\usepackage{dsfont}
\usepackage{color}
\usepackage{subcaption}
\graphicspath{ {./Figures/} }
\usepackage{lipsum}
\usepackage{etoolbox}
\usepackage[overload]{textcase}
\usepackage{makecell}
\usepackage{styles/slashbox}
\usepackage{cite}
\usepackage{enumitem}
\usepackage{array}
\usepackage{layouts}
\usepackage{wrapfig}
\usepackage{pifont}
\usepackage{multirow}
\usepackage{resizegather} 
\usepackage{stfloats}
\usepackage{siunitx,ragged2e}
\usepackage{placeins}

\newcommand{\concat}{,}

\let\tinymatrix\smallmatrix

\patchcmd{\tinymatrix}{\scriptstyle}{\scriptscriptstyle}{}{}
\patchcmd{\tinymatrix}{\scriptstyle}{\scriptscriptstyle}{}{}
\patchcmd{\tinymatrix}{\vcenter}{\vtop}{}{}
\patchcmd{\tinymatrix}{\bgroup}{\bgroup\scriptsize}{}{}

\def \emph{\textit}

\def\ie{\emph{i.e., }}

\def\etal{\emph{et al.}}

\def\bfA{\mathbf{A}}
\def\bfb{\mathbf{b}}

\def\bfc{\mathbf{c}}
\def\bfC{\mathbf{C}}

\def\bfe{\mathbf{e}}
\def\bfE{\mathbf{E}}

\def\bfh{\mathbf{h}}

\def\bfl{\mathbf{l}}

\def\bfr{\mathbf{r}}

\def\bfs{\mathbf{s}}

\def\bfu{\mathbf{u}}
\def\bfU{\mathbf{U}}
\def\bfv{\mathbf{v}}
\def\bfV{\mathbf{V}}

\def\bfW{\mathbf{W}}

\def\bfx{\mathbf{x}}

\def\bfz{\mathbf{z}}

\def\arraystretch{0.80}%

\begin{document}
\title{Exact Combinatorial Optimization with \\ Temporo-Attentional Graph Neural Networks}

\toctitle{Exact Combinatorial Optimization with Temporo-Attentional Graph Neural Networks}

\titlerunning{Combinatorial Optimization with \\ Temporo-Attentional GNNs}
%

\author{Mehdi Seyfi \and
Amin Banitalebi-Dehkordi \and
Zirui Zhou \and
Yong Zhang
}
\institute{Huawei Technologies Canada Co., Ltd.\\
yong.zhang3@huawei.com}

\tocauthor{Mehdi Seyfi, Amin Banitalebi-Dehkordi, Zirui Zhou, and Yong Zhang}

%

\maketitle              

\begin{abstract}
Combinatorial optimization finds an optimal solution within a discrete set of variables and constraints. The field has seen tremendous progress both in research and industry. With the success of deep learning in the past decade, a recent trend in combinatorial optimization has been to improve state-of-the-art combinatorial optimization solvers by replacing key heuristic components with machine learning (ML) models. In this paper, we investigate two essential aspects of machine learning algorithms for combinatorial optimization: temporal characteristics and attention. We argue that for the task of variable selection in the branch-and-bound (B\&B) algorithm, incorporating the temporal information as well as the bipartite graph attention improves the solver's performance. We support our claims with intuitions and numerical results over several standard datasets used in the literature and competitions.\footnote{Code is available at: \url{https://developer.huaweicloud.com/develop/aigallery/notebook/detail?id=047c6cf2-8463-40d7-b92f-7b2ca998e935}} 
\keywords{Combinatorial optimization \and Graph Neural Networks \and Temporal Attention \and Mixed Integer Linear Program.}
\end{abstract}

\section{Introduction}\label{sec_introduction}
Combinatorial optimization is the process of searching for extrema of an objective function with a discrete domain when the optimized variables satisfy some pre-defined constraints. Typical examples of such problems include: the Traveling Salesman Problem (TSP)\cite{flood1956traveling}, finding the Minimum Spanning Tree (MST) \cite{graham1985history}, and the Knapsack problem \cite{salkin1975knapsack}.

Combinatorial optimization is adopted in many critical applications affecting day-to-day lives. Examples include: daily electric grid power distribution \cite{knueven2020mixed,morais2010optimal}, airport flights scheduling \cite{bennell2017dynamic}, and
etc.
Due to the importance of such applications, there has been a tremendous amount of effort from both academia \cite{achterberg2009scip,vaz2009pswarm,fiala2013penlab} and industry \cite{manual1987ibm,bixby2007gurobi,gamrath2020scip} to build advanced and reliable solutions.

In general, many combinatorial optimization problems can be reduced to Mixed-Integer Linear Programs (MILPs) in which at least some of the variables in the feasible domain are integral and the objective function and constraints are linear \cite{hoffman2013integer}.
The existing MILP solutions, for the most part, are general-purpose one-size-fits-all products that target a variety of applications. However, in many applications, the data only changes slightly over time (e.g. daily electricity consumption in the same city should not change drastically day over day in a fixed network). These changes are hard to capture with hand-designed rules. This has motivated researchers to investigate the possibility of training machine learning models from the historical data, and use these models to help solve MILPs \cite{gasse2019exact,nair2020solving,gupta2020hybrid,khalil2017learning}. 

The standard well-established and exact approach to solving MILPs is the Branch and Bound (B\&B) algorithm \cite{land2010automatic}. 
Variable selection within B\&B is an essential step in which a fractional variable is selected in each LP relaxation iteration. The gold standard to perform variable selection is the Full Strong Branching (FSB) rule, which is unfortunately computationally expensive \cite{manual1987ibm}. Consequently, many algorithms try to propose a fast approximation of the FSB \cite{achterberg2005branching}.

In this paper, we focus on variable selection in the B\&B algorithm by mimicking the full strong branching via imitation learning \cite{hussein2017imitation}. Our intention is to use the statistical properties of the MILP data samples to train a neural network model that can learn to imitate the variable branching from the FSB algorithm with much less computational complexity. 
Building on the former attempts in the literature to tackle this problem \cite{khalil2016learning,gupta2020hybrid,alvarez2017machine,nair2020solving}, by adopting a bipartite graph representation for MILP problems, we propose to engage with variable selection via two novel contributions. First, we embed the MILP graph into representation vectors utilizing the Graph Attention Networks (GAT), which are the state-of-the-art structures for representation learning \cite{velickovic2017graph,brody2021attentive}. We argue that as opposed to the traditional Graph Convolutional Neural Network (GCNN) structures, our model allows for \emph{implicitly} assigning different gravity to nodes of the same neighborhood, enabling a surge in the model capacity. This would let our policy to capture information about the node embeddings that are more interesting to the expert solver (here FSB agent) to perform a branching action.
Second, by dividing the process of solving a MILP instance into consecutive episodes of a Markov decision process \cite{howard1960dynamic}, we propose to incorporate the temporal variations of representations associated to consecutive MILP episodes, into our smart branching scenario. To this end, we propose a Gated Recurrent Unit (GRU) to capture the temporal information concealed in the representation vectors associated with each episode of a MILP instance solution. We compare our results against the previous variable selection strategies in the literature and show that our method performs competitively compared to the existing branching mechanisms.

\section{Related Work}\label{sec_related_work}

Previous attempts to replace components of MILP solvers with machine learning models include:

\paragraph{Learning primal heuristics:} Authors in \cite{khalil2016learning, ding2020accelerating, shen2021learning} introduced methods to learn the primal heuristics; \ie methods with which a feasible but not necessarily optimal solution may be found. 
The task of learning primal heuristics is known as \emph{primal task} in the research community \cite{ml4co_gasse}.

\paragraph{Node selection:} Moreover, authors in \cite{he2014learning, song2018learning} studied the node selection. He \etal \cite{he2014learning} through imitation learning, learned a policy to select a candidate node with the optimal solution
in its sub-tree. Song \etal \cite{song2018learning} learned node selection and a good search policy via retrospective imitation learning, which is a self-correcting imitation learning algorithm by ruling out previous bad decisions. 

\paragraph{Learn to branch:} Authors in \cite{gasse2019exact, ml4co_gasse, ml4co_qu} trained neural networks that imitate the internal gold standard full strong branching mechanism for variable selection. Alvarez {\etal} proposed to approximate a branching function on hand-crafted features using Extremely Randomized Trees (ExtraTrees) \cite{geurts2006extremely}, a modified version of random forest \cite{breiman2001random}, which is based on an ensemble of regression trees. 
The authors in \cite{gasse2019exact} modeled the MILP-solving process by a Markov decision process \cite{howard1960dynamic}. At each state, the policy makes a decision on the optimal variable to branch on. They encode each MILP state by a GCNN and train their model with behavioral cloning \cite{pomerleau1991efficient} and a cross-entropy loss. This task is known as the \emph{dual task} in the research community \cite{ml4co_gasse}. 

In the Machine Learning for Combinatorial Optimization (ML4CO) competition \cite{ml4co_competition} held in 2021, the organizers challenged the participants in different tracks \ie the primal, the dual, and configuration tasks.
In the \emph{dual task} scenario which lies within the scope of this paper, the competition results revealed that the GCNN architecture used for branching can achieve a strong performance when combined with other techniques and tricks.
For example, the winner solution proposed Knowledge Inheriting Dataset Aggregation (KIDA) along with a Model Weight Averaging (MWA) mechanism \cite{ml4co_gasse} to be applied on the GCNN architecture. This solution used the GCNN model proposed by \cite{gasse2019exact} on an aggregated dataset using the techniques in \cite{ross2011reduction}. It trained multiple parent models and performed a greedy search to select the final model from the trained parent models and their children weight averaging models \cite{ml4co_gasse}. 
The runner-up team (EI-OROAS) in the same task also used the baseline GCNN \cite{gasse2019exact} and argued that the GCNN approach could be very effective if it was tuned and trained properly on the right kind of training samples \cite{ml4co_banitalebi}.
In a later approach, the authors in \cite{nair2020solving} combined a learned primal heuristic and a branching policy in the solver environment together in order to tackle more practical real-world problems. In particular, they proposed neural diving that learns primal heuristics and neural branching that learns a branching policy to achieve a better performance in terms of latency and accuracy. 

Although the GCNN-based methods set a good standard for selecting fractional variables in the B\&B algorithm, there is still room for developing lightweight models that can imitate the full strong branching rule more accurately. To this end, we investigate two essential aspects of machine learning algorithms for branching in combinatorial optimization: temporal characteristics and attention. We argue that for the task of variable selection in the branch-and-bound (B\&B) algorithm, incorporating the temporal information as well as the bipartite graph attention improves the solver's performance. 
\section{Background}\label{sec_background}
\paragraph{Preliminaries and definitions:}
A mixed-integer linear program is defined as:
\begin{equation}\label{eq_optim}
    \arg \min_{\bfx} \{\bfc^T\bfx|\bfA\bfx\leq\bfb, \bfl\leq \bfx\leq\bfu,\bfx\in\mathds{Z}^p\times\mathds{R}^{n-p}\},
\end{equation}
where $\bfc\in\mathds{R}^n$ denotes the coefficients of the linear objective, and $\bfA\in\mathds{R}^{m\times n}$ and $\bfb\in\mathds{R}^{m}$ respectively represent the coefficients and upper bounds of the linear constraints. There are $m$ linear constraints and $n$ variables where $p\leq n$ is the number of integer variables. $\bfl$ and $\bfu$ are both vectors in the $\mathds{R}^n$ space and are the lower and upper bound vectors on variables $\bfx=\left[x_1,\ldots,x_n\right]$.

A \emph{feasible solution} is a solution that satisfies all the constraints in \eqref{eq_optim}. A linear programming relaxation is when we relax the last constraint in \eqref{eq_optim}, \ie $\bfx \in\mathds{R}^n$. This will turn the MILP to a \emph{Linear Program} (LP) \cite{boyd2004convex}. The value of the objective function $\bfc^T\bfx$ with the LP solution is a lower bound to the original MILP. Any lower bound for the MILP is referred to as a \emph{dual bound}. The LP solution can be a feasible solution if it satisfies the integral constraints, \ie $\bfx\in\mathds{Z}^p\times\mathds{R}^{n-p}$.
The \emph{primal bound} is the objective value of a solution that is feasible for \eqref{eq_optim}, but not necessarily optimal. This could be an upper bound to the objective value of the MILP. Finally, the dual-primal gap is the gap between the dual bound and the primal bound.

\paragraph{The branch and bound algorithm:}\label{sec_B_and_B}
It is common in practice to solve the MILPs sequentially by building a search tree at each node with partial assignment of integer values to the variables, and use the information obtained at the node to converge to an optimal or a near-optimal solution \cite{land2010automatic,achterberg2013mixed,ding2020accelerating}. At each step, we choose a leaf node to branch from (choose a variable to branch). We solve the LP relaxation problem at this node where we constrain the previously branched variables to be fixed at their integer value. Therefore at each node, we relax $p-r$  variables where $r\leq p$ and make a decision on which variable to branch on. The LP solution at this node provides us with a lower bound to the objective value of the original MILP solution as well as any further child nodes down the road. If this lower bound is larger than the objective value of any known feasible solution then we can safely cutout this branch of the search tree as it is guaranteed that the child nodes of this particular node will provide us with a larger (worse) objective value. If the LP relaxation at this node is not larger than the objective value of a known feasible solution then we may decide to expand this node. We do that by branching on a variable from the remaining fractional variables at that node. Once a variable is selected, the tree ramifies into two branches, and two child nodes are added to the search tree. We divide the domain of the selected variable into two non-overlapping intervals. We choose the solution of the LP relaxation problem at the parent node for that particular variable as a reference. If $x_i^{lp}$ is the LP relaxation solution of the variable with index $i$ at the parent node, the non-overlapping domains of child nodes will be $x_i\geq  \lceil x_i^{lp}\rceil$ 
and $x_i \leq \lfloor x_i^{lp}\rfloor$, where $\lceil\cdot\rceil$ and 
$\lfloor\cdot\rfloor$ are the ceiling and floor operators, respectively. A new MILP 
the sample is generated from the MILP instance once branching on one variable is performed. The tree is updated and this procedure is resumed until convergence. LP is the backbone of the branch and bound algorithm. It is used for both finding the dual bounds at each node and deciding on the variable to branch on with the help of some primal heuristics. Practically the size of a search tree is in the exponential order with respect to the number of variables, therefore in some cases the search tree can be huge, and therefore time-consuming to traverse through.

\begin{figure*}[htb!]
\centering
\includegraphics[width=0.90\textwidth]{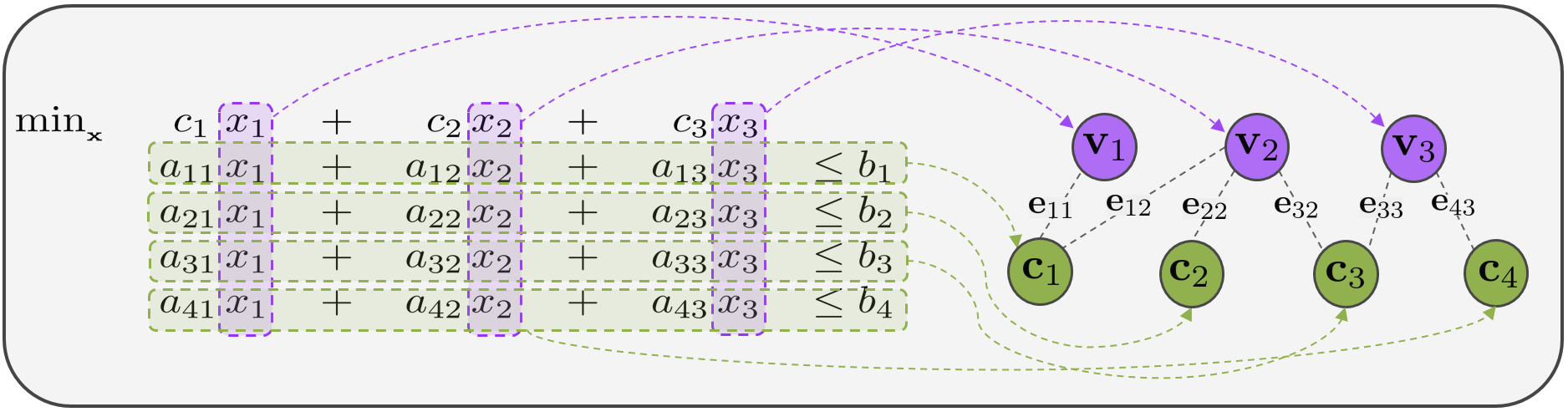}
\vspace{-6pt}
\caption{
An example representing a MILP instance of 3 variables and 4 constraints with a bipartite graph \cite{nair2020solving}. 
$\bfv_{j}\in \mathds{R}^{v}$, $\bfc_{i}\in \mathds{R}^{c}$, and $\mathbf{e}_{ij} \in \mathds{R}^{e}$ denote the $j^{\rm {th}}$ variable, $i^{\rm{th}}$ constraint, and the edge connecting the two.
In this example $a_{13} = a_{21}= a_{23}=a_{31}=a_{41}=a_{42}=0$; therefore, there is no connecting edge between their representing graph nodes. For brevity of illustration, we have ignored the time-dependent nature of the node/edge features.} 
\label{fig:graph}
\end{figure*}

\section{Methodology}\label{sec_method}
\begin{figure*}[htb!]
\centering
\begin{subfigure}{.35\textwidth}
   \includegraphics[width=1\linewidth]{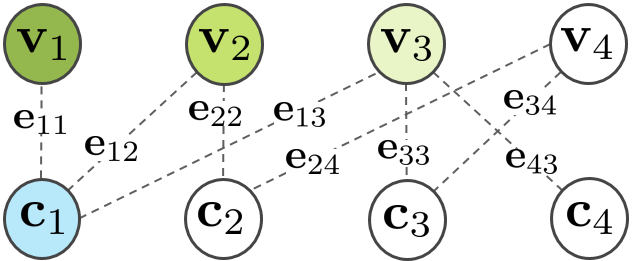}
   \label{fig:attention_cons} 
\vspace{-12pt}
\end{subfigure}\qquad\qquad
\begin{subfigure}{.35\textwidth}   \includegraphics[width=1\linewidth]{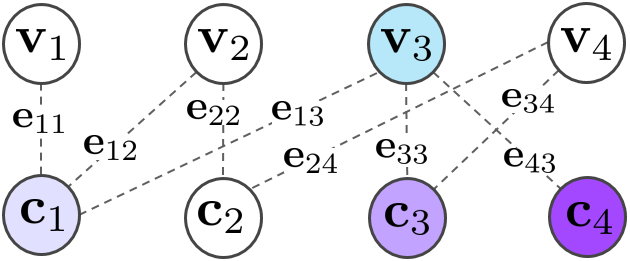}
\label{fig:attention_mechanism}
\vspace{-12pt}
\end{subfigure}
\vspace{-4pt}
\caption{(left) Intuitively, since $\bfv_1$ only appears in the first constraint; therefore $\bfc_1$ attends to $\bfv_1$ more, rather than $\bfv_3$ that participates in multiple constraints. It is mainly because the information about $\bfv_3$ flows in the graph not only via its connection to $\bfc_1$, but also via other connecting nodes to $\bfv_3$ other than $\bfc_1$. (right) With a similar intuition $\bfv_3$ attends to $\bfc_4$ the most and $\bfc_2$ the least. In both figures darker color means more attention.}
\label{fig:attention_layer}
\end{figure*}

In this section, we elaborate on the mechanics of our method for addressing variable selection in the B\&B algorithm within a time-limit $T$. As introduced by \cite{he2014learning} and later followed by \cite{gasse2019exact,gupta2020hybrid,nair2020solving} we can model the sequential selections made by the B\&B algorithm with a Markov decision process \cite{howard1960dynamic}. Letting the solver be the environment and the brancher the agent, \cite{gasse2019exact} denotes the solver state at the $t^{th}$ decision by $\bfs_t$, which contains information about the current dual bound, primal bound, the LP solution of each node, the current leaf node, etc. Let the action set $\mathcal{A}_t\subseteq\{1,\ldots,p\}$ be a set including the index of the fractional variables at the current LP relaxation node at the state $\bfs_t$. During a branching episode; the agent, based on the environment variables, and a selection policy $\pi_{\theta}(\cdot)$ with learning parameters $\theta$, takes an action $\tilde{a}_t\in\mathcal{A}_t$ which points to the index of a \emph{desirably} optimal fractional variable to branch on; performs the branching-and-bounding as stated in Sec.~\ref{sec_B_and_B} and moves to the next state $\bfs_{t+1}$. 
The authors in \cite{gasse2019exact,gupta2020hybrid,nair2020solving,rl4co} encode each state $\bfs_t$ of the B\&B Markov process at time slot $t$ as a bipartite graph $\mathcal{G}$ with node and edge features $(\mathcal{G},\bfC_t, \bfV_t,\bfE_t)$. At the current node's LP relaxation, each row in the feature matrices $\bfC_t\in\mathds{R}^{m\times c}$ and $\bfV_t \in \mathds{R}^{n\times v}$ represents a row and a column of the MILP instance at the state $\bfs_t$, respectively (ref to Fig.~\ref{fig:graph}).
In this setting, $\bfv_{j,t}$ and $\bfc_{i,t}$ refer to the $j^{\rm th}$ and the $i^{\rm th}$ rows from $\bfC_t$ and $\bfV_t$, respectively. Besides,  
node $\bfc_{i,t}$ is connected to the node $\bfv_{j,t}$ via the edge $\bfe_{ij,t}\in\mathds{R}^{e}$ if and only if $a_{ij}\neq0$ (ref. Fig.~\ref{fig:graph}).
Subsequently, the sparse feature tensor $\bfE_t\in\mathds{R}^{m\times n \times e}$ concatenates all $\bfe_{ij,t}$ features. $c$, $v$, and $e$ represent the dimensions of the feature vectors for constraints, variables, and edges, respectively.
 The aforementioned feature vectors are obtained by extracting some hand-crafted features from the solver environment. The authors in \cite{gasse2019exact,hutter2011sequential} studied and proposed engineering such features. We leverage the same set of features proposed in \cite{gasse2019exact} in our work.
In the following sub-sections we elaborate on our methodology and the components of our neural branching mechanism to imitate the FSB in the solver environment.
\paragraph{Embedding layers:}
To increase the modeling capacity and also to be able to manipulate the node interactions with our proposed neural architecture, following \cite{gasse2019exact,gupta2020hybrid,ml4co_banitalebi} we use embedding layers to map each node and edge to space $\mathds{R}^d$. For brevity and simplicity of notation, in the forthcoming sections, we assume that the embedding layers are already applied to $(\mathcal{G},\bfC_t, \bfV_t,\bfE_t)$ and therefore, $(\bfc_{i,t}, \bfv_{j,t}, \bfe_{ij,t})\in\mathds{R}^{d\times d \times d} , \forall(i,j,t): 1\leq i \leq m,1\leq j\leq  n, 0\leq t\leq T$.
\paragraph{Attention mechanism:} 
Neighborhood normalization, in many cases, is known to be useful for improving the AGGREGATE operator in the Message Passing Networks (MPN) \cite{kipf2016semi}. The intuition behind this normalization is that higher-degree neighbors might be bearing more generic and less precise information; therefore, the model should put less stress on such nodes. On the other hand, in some cases, normalization may lead to loss of information by removing key structural information from the graph nodes. Specifically, the embedding learned from nodes with different degrees might be indistinguishable \cite{Hamilton_grl}.    
Intuitively, some kind of node normalization for a graph representation of a MILP instance may be justifiable. The variables participating in many constraints
might be less information-bearing than the ones engaging in only a few (ref. Fig.~\ref{fig:attention_layer}). At the same time, by normalizing the node degrees, we might be removing some structural information from the graph representation $(\mathcal{G},\bfC_t, \bfV_t,\bfE_t)$. Therefore, we propose to use an attention mechanism to extract the information associated with the interplay between the nodes. By using attention, we give the model the freedom to prioritize each node according to its neighborhood structure and embedding features. Doing so will let the model decide how much participation a node should have in the final decision-making policy.

Considering the bipartite nature of $(\mathcal{G},\bfC_t, \bfV_t,\bfE_t)$, we use a pair of back-to-back attention structures to encode the node interactions. Each constraint node $\bfc_{i,t}$ attends to its neighborhood $\mathcal{N}_i$ in the first round via an attention structure with number of $H$ attention heads:
\begin{equation}\label{eq:cons_attention}
\bfc_{i,t} = \frac1{H}\sum_{h=1}^{H}\left(\alpha_{ii}^{(h)}\mathbf{\Theta}^{(h)}_c\bfc_{i,t} + \sum_{j\in \mathcal{N}_i}\alpha_{ij}^{(h)}\mathbf{\Theta}^{(h)}_v\bfv_{j,t}\right),
\end{equation}
with learnable weights $\mathbf{\Theta}^{(h)}_c,\mathbf{\Theta}^{(h)}_v\in\mathds{R}^{d'\times d}$ and LeakyRelu \cite{xu2015empirical} being the activation function. The updated constraint embeddings are averaged across multiple attention heads using attention weights \cite{brody2021attentive}:
\begin{equation}\label{eq:attention_alpha}
\alpha_{ij}^{(h)} =
\frac{
\exp\left(\mathbf{a}_c^{{(h)}^\top}\mathrm{LeakyReLU}\left(
[\mathbf{\Theta}^{(h)}_c\bfc_{i,t} \, \concat \, \mathbf{\Theta}^{(h)}_v\bfv_{k,t} \, \concat \, \mathbf{\Theta}^{(h)}_e\mathbf{e}_{ik,t}]
\right)\right)}
{\sum_{k \in \mathcal{N}_i \cup \{ i \}}
\exp\left(\mathbf{a}_c^{{(h)}^\top}\mathrm{LeakyReLU}\left(
[\mathbf{\Theta}^{(h)}_c\bfc_{i,t} \, \concat \, \mathbf{\Theta}^{(h)}_v\bfv_{k,t} \, \concat \, \mathbf{\Theta}^{(h)}_e\mathbf{e}_{ik,t}]
\right)\right)},
\end{equation}
where $\mathbf{\Theta}^{(h)}_e\in\mathds{R}^{d'\times d}$ is a learnable weight.
The attention coefficients vector $\mathbf{a}_c^{(h)}\in \mathds{R}^{3d'}$, is automatically learned to encode both feature level and structure level information flow in the graph and ``$\concat$'' denotes vector concatenation.
Similarly, the variable nodes are encoded via:
\begin{equation}\label{eq:var_attention}
\bfv_{j,t} = \frac1{H}\sum_{h=1}^H\left(\beta_{jj}^{(h)} \Psi^{(h)}_v\bfv_{j,t} + \sum_{i\in N_j} \beta_{ji}^{(h)}\Psi^{(h)}_c\bfc_{i,t}\right),
\end{equation}
with learnable weights $\Psi^{(h)}_v \in \mathds{R}^{d \times d}$, $\Psi^{(h)}_c\in\mathds{R}^{d\times d'}$, and: 
\begin{equation}\label{eq:attention_beta}
\beta_{ji}^{(h)} =
\frac{
\exp\left(\mathbf{a}_v^{{(h)}^{\top}}\mathrm{LeakyReLU}\left(
[\mathbf{\Psi}^{(h)}_v\bfv_{j,t} \, \concat \, \mathbf{\Psi}^{(h)}_c\bfc_{i,t} \, \concat \, \mathbf{\Psi}^{(h)}_e\mathbf{e}_{ji,t}]
\right)\right)}
{\sum_{k \in \mathcal{N}_j \cup \{ j \}}
\exp\left(\mathbf{a}_v^{{(h)}^{\top}}\mathrm{LeakyReLU}\left(
[\mathbf{\Psi}^{(h)}_v\bfv_{j,t} \, \concat \, \mathbf{\Psi}^{(h)}_c\bfc_{k,t} \, \concat \, \mathbf{\Psi}^{(h)}_e\mathbf{e}_{jk,t}]
\right)\right)},
\end{equation}
where $\mathbf{\Psi}^{(h)}_e\in \mathds{R}^{d\times d}$ and $\mathbf{a}_v^{(h)}\in \mathds{R}^{3d}$
are learnable weights and attention coefficients vector. 
The constraint feature nodes in \eqref{eq:var_attention} and \eqref{eq:attention_beta} are replaced by their updated value in \eqref{eq:cons_attention}.

Feature nodes $\bfv_{i,t}$ encode the LP relaxation state of each variable in the current node $\forall i \in [0, n]$. These encoded representations hold information about the graph structure and node embeddings of the MILP instance at the state $\bfs_t$. 
\paragraph{Temporal encoding:}
After the $t^{\rm{th}}$ branching episode the solver state $\bfs_t$ which was represented by the graph $\left(\mathcal{G},\bfV_t,\bfC_t,\bfE_t\right)$ is further
encoded to a set of variable features $\bfv_{i,t}, \forall i\in\{1,\ldots,n\}$ via
passing the bipartite graph through a back-to-back attention module. This graph 
representation of the solver state; however, encodes only the current B\&B tree state and lacks the temporal information about the past node/edge features that have led the graph representations to the current state. To better imitate the agent in the solver environment, monitoring the temporal variations of the encoded graph carries critical information about the temporal variations in the node/edge embeddings and their relative temporo-structural interplay. To this end, we can inject crucial information about the variation of the features associated to the B\&B tree, and what sequential features have led the tree to the current status, into our model. To capture this temporal interaction between the graph nodes/edges we utilize a single-layer GRU recurrent neural network (RNN) to a sequence of $L$ consecutive variable embeddings $\bfv_{i,t} ,  \forall t \in \{t-L+1, \ldots, t\}$. Specifically for each variable node $\bfv_{i,t}$ in the input sequence and $t \in \{t-L+1, \ldots, t\}$, the model computes:
\begin{align}\label{eq_gru}
&\bfz_{i,t} = \sigma_g(\bfW_{z} \bfv_{i,t} + \bfU_{z} \bfh_{t-1} + \bfb_z), \nonumber\\
&\bfr_{i,t} = \sigma_g(\bfW_{r} \bfv_{i,t} + \bfU_{r} \bfh_{i,t-1} + \bfb_r), \nonumber\\
&\hat{\bfh}_t = \phi_h(\bfW_{h} \bfv_{i,t} + \bfU_{h} (\bfr_{i,t} \odot \bfh_{i,t-1}) + \bfb_h), \nonumber\\
&\bfh_{i,t} =  (\mathbf{1} - \bfz_{i,t}) \odot \bfh_{i,t-1} + \bfz_{i,t} \odot \hat{\bfh}_t,
\end{align}
where $\odot$ is the Hadamard product operator,
$\bfh_t\in\mathds{R}^{d''}$ is the  output vector,
$\hat{\bfh}_t\in\mathds{R}^{d''}$ is the candidate activation vector, $\bfz_t\in\mathds{R}^{d''}$ is the update gate vector, and $\bfr_t\in\mathds{R}^{d''}$ is the reset gate vector. 
$\bfW, \bfU \in \mathds{R}^{d''\times d}$, and $\bfb\in \mathds{R}^{d''}$ are GRU parameter matrices/vector, and
$\sigma_g$  and
$\phi_h$ are sigmoid and hyperbolic tangent activation functions.
Finally our branching policy models variable selection via:
\begin{equation}
    \pi_\theta(\tilde{a}_t|\bfs_t,\ldots,\bfs_{t+1-L} ) =\arg \max_{i}\frac{ \exp\left(F_\mathcal{V}\left(\bfh_{i,t}\right)\right)}{\sum_{j=1}^n\exp\left(F_\mathcal{V}\left(\bfh_{j,t}\right)\right)},
\end{equation}
where $F_{\mathcal{V}}: \mathds{R}^{ d''}\rightarrow \mathds{R}$ is a multi-layer perceptron. In the training time the model weights are updated via a gradient decent algorithm by minimizing the loss function: 
\begin{equation}
\mathcal{L}(\theta) = -\frac1L\sum_{l=t-L+1}^{t}\log\left(\pi_\theta(\tilde{a}_l|\bfs_l,\ldots,\bfs_{l+1-L} )\right).    
\end{equation}

%


\section{Experiments}\label{sec:experiments}
In this section, we present experiments and ablations to validate our theoretical propositions. We use SCIP 7.0 optimization suite \cite{GamrathEtal2020ZR} as the backend solver, along with the Ecole \cite{prouvost2020ecole} library to run experiments on a V100 GPU card with 32GB memory. For both generating the training set and solving the MILP instances we use a solver time-limit of 3600 seconds unless otherwise stated. All results are reported by averaging 5 separate runs with different seeds in the inference time. More details and ablation studies are provided in the \emph{appendix}.
\paragraph{Datasets:}\label{sec:setup}
We evaluate our method on six different datasets that cover a good range of variations in terms of difficulty among the available MILP benchmarks. These datasets include: Set Covering (SC) \cite{gasse2019exact}, Combinatorial Auctions (CA) \cite{gasse2019exact,leyton2000towards}, Capacitated Facility Locations (CFL) \cite{cornuejols1991comparison}, Maximum Independent Set (MIS) \cite{gasse2019exact,bergman2016decision}, work load appointments/Load Balancing (LB) \cite{ml4co_gasse}, and Maritime Inventory Routing (MIR)\cite{papageorgiou2014mirplib}. Details on how each benchmark is created are provided in the \textit{appendix}.
\paragraph{\textit{Remark}:} 
It is worth noting that, since solvers rely heavily on the underlying hardware of the testing machine (CPU, memory, GPU, etc.), a truly fair evaluation is only achieved when all baseline methods are run on the same machine with the same set of MILP instance; for this, we either trained the baselines from scratch on the same MILP samples or evaluated the checkpoint provided by the authors on the same MILP instances in our environment.
\paragraph{Baselines}\label{sec:baseline}
We compare our results with SCIP's internal branching: FSB, reliability pseudocost branching (RPB) \cite{achterberg2005branching}, and the pseudo cost branching rule (PB). Additionally, for the first 4 benchmarks, we compare our results with the GCNN approach of Gasse \etal ~\cite{gasse2019exact}, LambdaMART \cite{burges2010ranknet}, SVMRank \cite{khalil2016learning} and finally the ExtraTrees method proposed by \cite{geurts2006extremely}. For this, we used the code base provided by \cite{gupta2020hybrid} in our environment. For the last two benchmarks we compare our results with the internal branching rules of SCIP and also the method proposed by \cite{ml4co_Nuri} and EI-OROAS from the ML4CO competition \cite{ml4co_gasse}.
\paragraph{Training}\label{sec:training}
For training our temporo-attentional branching policy, we run a training data collection phase in which the instances are solved with a time-limit of 3600 seconds using the FSB rule from SCIP as our expert agent. For each benchmark, we generate 160k samples from the training set instances for all the benchmarks except for maritime inventory routing dataset that we generated only 5.7k MILP samples due to lack of enough training MILP instances. In particular we record the states of the first $L$ \emph{consecutive} episodes of each MILP instance in the form of bipartite graph representations along with the branching choices associated to each episode.
The agents are then trained with the collected datasets. Further details of the training procedure is given in the \emph{appendix}.
\paragraph{Metrics of performance}
For the first 4 benchmarks we use the same evaluation metrics as in \cite{gasse2019exact, gupta2020hybrid}. Specifically, we report:
Time: the 1-shifted geometric mean of solving time across the Easy, Medium, and Hard segments of each benchmark. 
Node: 1-shifted geometric mean of B\&B node count of the instances solved by each strategy.
Win: number of times each branching agent wins the other strategies based on the solving time across multiple validation runs.
\begin{figure*}[htb]
\centering
\begin{subfigure}[b]{0.70\textwidth}
   \includegraphics[width=.49\linewidth]{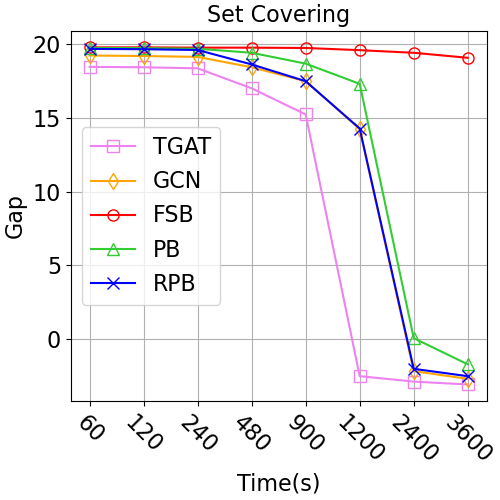}
   \label{fig:Ng1}
   \includegraphics[width=.49\linewidth]{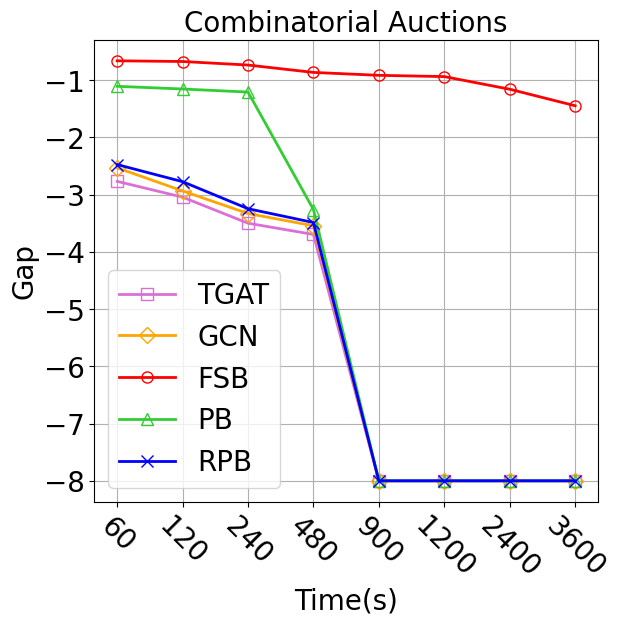}
   \label{fig:Ng2}
   \includegraphics[width=.49\linewidth]{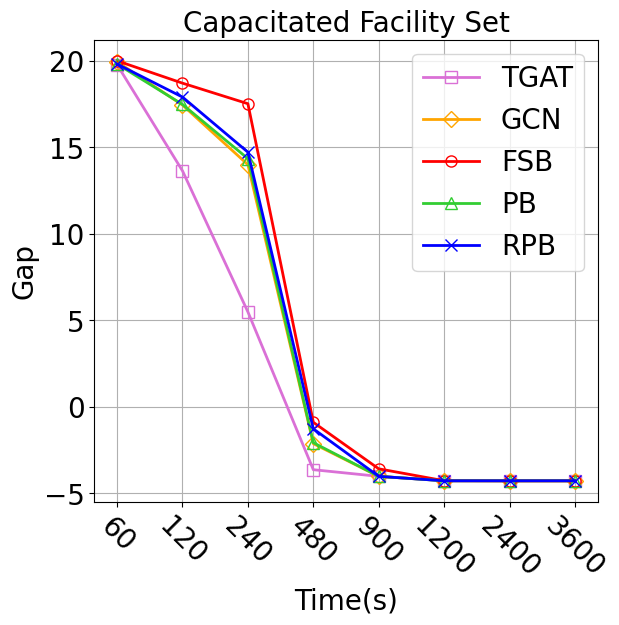}
   \label{fig:Ng3}
   \includegraphics[width=.49\linewidth]{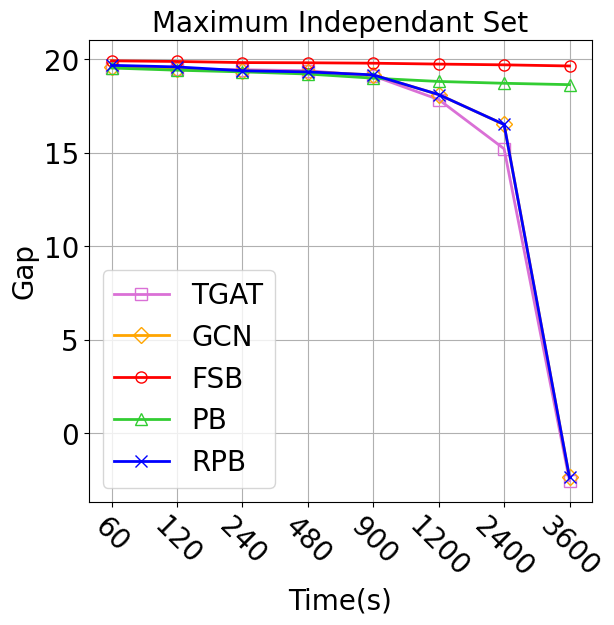}
   \label{fig:Ng4}
\end{subfigure}
\vspace{-8pt}
\caption{Average dual-primal gap (logarithmic scale) vs solving time-limit (seconds).}
\label{fig:dual_primal_gap}
\end{figure*}

\paragraph{\textit{Remark:}} It is worth noting that the metrics mentioned above, each one alone, doesn't fully capture the solvers performance; since for each MILP instance the rate with which the policy approaches to the optimal solution is important. In other words, a good solution should be able to reduce the gap to the optimal value in a short amount of time. Therefore, it makes sense to include the rate with which the gap is reduced in the evaluation metric. To this end, the ML4CO competition \cite{ml4co_gasse} incorporated a `reward' metric to address this for the last two benchmarks. This metric is defined as:
\begin{equation}
\mathcal{R} = \int_{t=0}^{T} \mathbf{z}^\star_t\ \mathrm{d}t	- T\mathbf{c}^\top\mathbf{x}^\star,
\end{equation}
where $\bfz^\star_t$ is the best dual bound at time $t$, $\bfx^\star$ is the optimal solution and $T$ is the time-limit. The reward, within a time-limit of $T$, is maximized if the gap between the optimal solution and the dual bound is decreased with a higher rate during consecutive episodes of the branching process.
\paragraph{Results and discussions:}
Table \ref{tab:main_results} shows the results on the first 4 benchmark datasets compared to the baselines in three segments of the datasets \ie Easy, Medium, and Hard instances, where the GAT structure is parameterized with $(d,H)$ where $d=d'=d''$, and $H$ is the number of attention heads. Consequently, the temporo-attentional (TGAT) method is parameterized with $(d,H,L)$ with $L$ being the GRU sequence length.  Ablation study on the hyper-parameters is provided in the \textit{appendix}. The \emph{Node} and \emph{Time} metrics are reported when applying the policies on 20 test instances per dataset per difficulty segment, averaged over 5 runs(total 100 instances). As it can be seen our method outperforms the other baselines in terms of the evaluation metrics Wins, and Time for the set covering, capacitated facility locations, and maximum independent set benchmarks.
In all the cases our model outperforms the baseline GCNN method \cite{gasse2019exact}. 
Amongst other baselines LambdaMart performs better in Easy evaluation instances; however, its performance degrades in Hard problems. Figure \ref{fig:dual_primal_gap}, shows the dual-primal gap \cite{nair2020solving} across the branching policies. As observed, our methods perform better than the other internal branching rules, as well as the GCNN baseline in closing the gap between the dual bound and the primal bound during a given solving time-limit. Among the internal branching rules FSB is the slowest and RPB is the fastest in closing the dual-primal gap.

\begin{figure}
\centering
\vspace{-10pt}
   \includegraphics[width=0.55\linewidth]{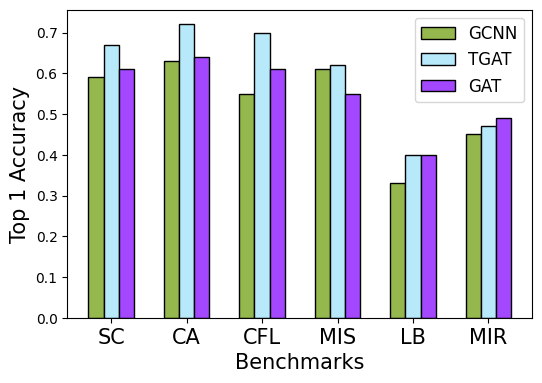}
\vspace{-10pt}
\caption{Top-1 accuracy for different branching policies.}
\vspace{-12pt}
\label{fig:top1_acc}
\end{figure}

\paragraph{Ablation on TGAT vs GAT:}
To evaluate the effect of incorporating the temporal characteristics of the variable embeddings we evaluate the GAT-only agent by bypassing the GRU structure in our model. 
Figure \ref{fig:top1_acc}, shows the top-1 validation accuracy of our proposed methods vs GCNN for different benchmarks. For all the datasets TGAT outperforms both GCNN and GAT in terms of imitating the FSB branching expert. 
Additionally, Table \ref{tab:main_results} shows that TGAT outperforms the GAT agent except for the combinatorial auctions dataset. We argue that since this benchmark has relatively smaller MILP instances, adding a GRU structure to the model increases the policy complexity and thus the inference time. According to our metrics, the policy that can close the dual gap (reach the optimal solution) in a shorter solving time wins. For small and easy MILP instances a lightweight policy with less branching accuracy may win if it can solve B\&B nodes at a higher rate. Our GAT version of the proposed algorithm, however, still outperforms other branching baselines. 
Although adding to the sequence length $L$, helps the TGAT policy to branch more accurately, it adds to the model complexity and increases the inference time, which as discussed above may degrade the branching performance; therefore, the sequence length should be tuned according to the MILP dataset complexity (More details in the \emph{appendix}).

\paragraph{Dual integral reward:} Following the metric proposed in \cite{ml4co_gasse} we report the dual integral rewards for the load balancing and the maritime inventory routing benchmarks in Tables \ref{tab:load_balancing} and \ref{tab:rewad_anonymous}, respectively.

\newcommand{\STAB}[1]{\begin{tabular}{@{}c@{}}#1\end{tabular}}

\begingroup
\setlength{\tabcolsep}{0.8pt} 
\renewcommand{\arraystretch}{1.01} 
\begin{table*}[]
\centering
    \caption{Evaluation of branching strategies for sets of easy, medium, and hard MILP instances in terms of time, wins, and nodes metrics along with the standard deviation across runs. The $^\dagger$ superscript indicates our methods.}
    \label{tab:main_results}
\vspace{1pt}
\fontsize{5.6}{7.5}\selectfont
{
    \centering
    \begin{tabular}{c}
        \begin{tabular}{l|ccccccccc}
        & \multicolumn{3}{c}{Easy}&\multicolumn{3}{c}{Medium}&\multicolumn{3}{c}{Hard}\\
        Model&Time$\downarrow$&Wins$\uparrow$&Nodes$\downarrow$&Time$\downarrow$&Wins$\uparrow$&Nodes$\downarrow$&Time$\downarrow$&Wins$\uparrow$&Nodes$\downarrow$\\
         \Xhline{2\arrayrulewidth}
\\
\multicolumn{10}{c}{Set Covering} \\
FSB&21.2$\pm$6.5&0/100&19$\pm$0&488.6$\pm$145.4&0/100&183$\pm$4&3,601$\pm$58&0/100&n/a\\
\cline{1-10}
PB&9.1$\pm$2.6&5/100&286$\pm$5&75.5$\pm$14.6&0/100&2,532$\pm$32&2,351$\pm$52&0/100&83,329$\pm$1453\\
RPB&11.9$\pm$5.7&0/100&\textbf{56}$\pm$1&74.8$\pm$15.7&0/100&1,892$\pm$47&1,858$\pm$20&0/100&49,321$\pm$1065\\
SVMRANK\cite{khalil2016learning}&10.8$\pm$3.0&0/100&170$\pm$1&91.4$\pm$2.6&0/100&1,982$\pm$44&2,719$\pm$29&0/100&42,913$\pm$1085\\
L-MART\cite{burges2010ranknet}&9.5$\pm$4.6&4/100&168$\pm$28&77.8$\pm$13.1&0/100&2,005$\pm$55&2,432$\pm$40&0/100&45,823$\pm$991\\
EX-TREES\cite{geurts2006extremely}&12.3$\pm$1.3&0/100&174$\pm$4&122.1$\pm$3.3&0/100&2,281$\pm$58&3,033$\pm$64&0/100&60,123$\pm$1080\\
GCNN\cite{gasse2019exact}&8.3$\pm$1.6&10/100&140$\pm$3&65.5$\pm$1.1&8/100&1,586$\pm$45&1,745$\pm$39&0/100&31,234$\pm$487\\
FILM&8.7$\pm$2.5&2/100&145$\pm$3&67.2$\pm$1.2&3/100&1,626$\pm$16&1,995$\pm$38&0/100&37,234$\pm$820\\
\cline{1-10}
GAT$^{\dagger}$(32,2)&9.8$\pm$2.7&9/100&141$\pm$8&57.4$\pm$14.8&11/100&1,467$\pm$19&1,574$\pm$47&8/100&30,812$\pm$811\\
TGAT$^{\dagger}$(32,2,4)&\textbf{6.8}$\pm$0.20&\textbf{70}/100&126$\pm$9&\textbf{45.6}$\pm$1.2&\textbf{78}/100&\textbf{1,332}$\pm$25&\textbf{1,376}$\pm$24&\textbf{92}/100&\textbf{29,452}$\pm$313\\
\Xhline{1\arrayrulewidth}
\\      
\multicolumn{10}{c}{Combinatorial Auctions}\\

FSB&5.8$\pm$2.2&0/100&7$\pm$0&101.0$\pm$22.0&0/100&79$\pm$1&2,034$\pm$25&0/100&437$\pm$5\\
\cline{1-10}
PB&3.1$\pm$1.0&3/100&271$\pm$4&22.2$\pm$2.9&1/100&2,844$\pm$37&297$\pm$5&0/100&14,130$\pm$206\\
RPB&4.2$\pm$1.2&0/100&\textbf{12}$\pm$0&21.2$\pm$2.2&2/100&717$\pm$10&161$\pm$3&0/100&5,664$\pm$71\\
SVMRANK\cite{khalil2016learning}&3.4$\pm$1.3&2/100&79$\pm$1&26.4$\pm$2.8&0/100&911$\pm$12&442$\pm$7&0/100&6,964$\pm$137\\
L-MART\cite{burges2010ranknet}&2.9$\pm$0.8&40/100&81$\pm$1&16.0$\pm$4.4&63/100&919$\pm$17&241$\pm$4&0/100&7,179$\pm$135\\
EX-TREES\cite{geurts2006extremely}&3.9$\pm$1.9&1/100&89$\pm$1&37.6$\pm$10.6&0/100&1,022$\pm$15&908$\pm$15&0/100&11,387$\pm$207\\
GCNN\cite{gasse2019exact}&3.3$\pm$1.6&2/100&78$\pm$0&24.6$\pm$3.5&0/100&708$\pm$10&143$\pm$1&2/100&5,929$\pm$94\\
FILM&3.7$\pm$1.3&0/100&74$\pm$0&30.2$\pm$7.8&0/100&705$\pm$13&265$\pm$2&0/100&6,421$\pm$66\\
\cline{1-10}
GAT$^{\dagger}$(32,2)&\textbf{2.7}$\pm$1.3&\textbf{42}/100&67$\pm$0&\textbf{17.3}$\pm$4.4&\textbf{34}/100&\textbf{675}$\pm$7&\textbf{89}$\pm$1&\textbf{95}/100&\textbf{5,635}$\pm$74\\
TGAT$^{\dagger}$(32,2,2)&3.1$\pm$1.1&10/100&76$\pm$0&22.1$\pm$2.8&2/100&690$\pm$7&142$\pm$1&3/100&5,900$\pm$83\\
 \Xhline{1\arrayrulewidth}
\\
\multicolumn{10}{c}{Capacitated Facility Location}\\
FSB&33.2$\pm$8.4&0/100&16$\pm$0&229.4$\pm$67.8&0/100&82$\pm$0&784$\pm$11&0/100&61$\pm$1\\
\cline{1-10}
PB&25.1$\pm$10.0&0/100&157$\pm$1&143.5$\pm$31.8&0/100&411$\pm$8&544$\pm$7&0/100&408$\pm$5\\
RPB&28.9$\pm$9.5&0/100&\textbf{24}$\pm$0&169.9$\pm$37.7&0/100&\textbf{131}$\pm$2&607$\pm$6&0/100&\textbf{121}$\pm$4\\
SVMRANK\cite{khalil2016learning}&26.4$\pm$10.5&1/100&125$\pm$1&136.3$\pm$39.1&1/100&348$\pm$6&536$\pm$6&0/100&340$\pm$6\\
L-MART\cite{burges2010ranknet}&27.8$\pm$9.7&0/100&121$\pm$1&141.6$\pm$21.2&0/100&355$\pm$4&550$\pm$10&0/100&332$\pm$6\\
EX-TREES\cite{geurts2006extremely}&33.9$\pm$8.7&0/100&143$\pm$1&194.1$\pm$21.7&0/100&412$\pm$7&758$\pm$11&0/100&399$\pm$6\\
GCNN\cite{gasse2019exact}&24.6$\pm$10.9&5/100&112$\pm$1&130.2$\pm$38.3&2/100&345$\pm$5&519$\pm$7&1/100&348$\pm$6\\
FILM&22.1$\pm$8.1&6/100&110$\pm$2&127.1$\pm$25.9&3/100&361$\pm$7&501$\pm$9&1/100&340$\pm$5\\
\cline{1-10}
GAT$^{\dagger}$(32,2,2)&20.4$\pm$9.0&18/100&107$\pm$2&123.3$\pm$14.2&15/100&329$\pm$5&432$\pm$8&8/100&328$\pm$5\\
TGAT$^{\dagger}$(32,2,2)&\textbf{17.9}$\pm$8.7&\textbf{70}/100&99$\pm$1&\textbf{110.4}$\pm$22.6&\textbf{79}/100&304$\pm$3&\textbf{349}$\pm$5&\textbf{90}/100&301$\pm$3\\
\Xhline{1\arrayrulewidth}

\\
\multicolumn{10}{c}{Maximum Independent Set}\\
FSB&28.7$\pm$13.1&0/100&9$\pm$0&1,550.4$\pm$341.9&0/100&41$\pm$0&3,601$\pm$55&0/100&n/a
\\
\cline{1-10}
PB&11.1$\pm$3.7&0/100&6194$\pm$79&834.8$\pm$83.7&0/100&1,889$\pm$20&3,483$\pm$34&0/100&51,230$\pm$677\\
RPB&11.8$\pm$5.7&0/100&\textbf{29}$\pm$0&143.8$\pm$27.4&1/100&742$\pm$9&2,210$\pm$32&0/100&2,742$\pm$30\\
SVMRANK\cite{khalil2016learning}&13.5$\pm$5.1&0/100&59$\pm$1&273.9$\pm$75.4&0/100&\textbf{583}$\pm$9&3,036$\pm$42&0/100&6,852$\pm$127\\
L-MART\cite{burges2010ranknet}&9.8$\pm$2.7&7/100&61$\pm$1&190.8$\pm$20.8&0/100&795$\pm$9&3,071$\pm$32&0/100&9,171$\pm$132\\
EX-TREES\cite{geurts2006extremely}&13.1$\pm$4.3&0/100&81$\pm$1&1,730.4$\pm$203.4&0/100&5,123$\pm$85&3,601$\pm$64&0/100&40,562$\pm$674\\
GCNN\cite{gasse2019exact}&11.6$\pm$5.5&0/100&51$\pm$0&144.2$\pm$32.9&1/100&1,870$\pm$23&2,192$\pm$40&7/100&2,839$\pm$51\\
FILM&17.5$\pm$7.0&0/100&67$\pm$0&230.2$\pm$23.7&0/100&981$\pm$10&3,142$\pm$43&0/100&41,234$\pm$760\\
\cline{1-10}
GAT$^{\dagger}$(32,2,4) &8.8$\pm$3.6&23/100&47$\pm$0&137.0$\pm$39.0&3/100&1,611$\pm$16&2,171$\pm$25&9/100&\textbf{2,736}$\pm$49\\
TGAT$^{\dagger}$(32,2,4)&\textbf{8.5}$\pm$3.4&\textbf{70}/100&44$\pm$0&\textbf{96.1}$\pm$14.7&\textbf{95}/100&1,464$\pm$14&\textbf{2,126}$\pm$23&\textbf{84}/100&2,753$\pm$53\\
        \Xhline{1\arrayrulewidth}
        \end{tabular}
    \end{tabular}
}
\end{table*}
\endgroup

\begingroup
\setlength{\tabcolsep}{2.4pt} 
\renewcommand{\arraystretch}{1} 
\begin{table*}[]
\centering
    \caption{Dual Integral Reward for the load balancing Dataset.}
\vspace{1pt}
\scriptsize
{
    \begin{tabular}{c}
         \Xhline{2\arrayrulewidth}
        \begin{tabular}{l|cccccccc}
            {Method | Time}& 60s&120s&240s&480s&900s&1200s&2400s&3600s\\    
             \Xhline{1\arrayrulewidth}
  FSB&           \textbf{42236}&	 \textbf{84200}&	168,058&	335,839&	629,429& 839,126&		1,678,126&	2,517,404\\
 \Xhline{1\arrayrulewidth}

PB&            41951&	  83933&	168,003&	336,290&	630,968& 841,510&		1,683,792&	2,526,330\\
GCNN\cite{gasse2019exact}&         41,960&	83,944&	 167,997&	336,272&	630,889&	841,383&	1,683,626&	2,526,162\\
EI-OROAS\cite{ml4co_banitalebi}&     41,938&	83,921&	168,066&	336,539&	631,460& 842,240&		1,685,519&	2,529,120\\
Nuri\cite{ml4co_Nuri}&           41,951&       83,934&   168,034&         336,299&	630,989& 841,546&		1,683,857&	2,527,290\\
\Xhline{1\arrayrulewidth}

GAT$^{\dagger}(32,3)$& 41,952&	83,949&	168,068&	336,408&	631,120&	841,685&	1,684,149&	2,527,838\\
TGAT$^{\dagger}(32,3,4)$& 41,952&	83,950&	\textbf{168,123}&	\textbf{336,654}&	\textbf{631,675}& \textbf{842,527}&		\textbf{1,686,093}&	\textbf{2,529,981}
\end{tabular}
    \end{tabular}
    \label{tab:load_balancing}
}
\end{table*}
\endgroup

We evaluate the results on the same test set used by \cite{ml4co_gasse} in the ML4CO challenge. We compare our results with the SCIP's internal branching rules, the GCNN \cite{gasse2019exact}, team Nuri \cite{ml4co_Nuri}, and team EI-OROAS \cite{ml4co_banitalebi} of the competition (Nuri \& El-OROAS results are reproduced in our environment using the checkpoints provided by the authors).
We observe that FSB outperforms other policies in small time-limits. For such time-limits, none of the policies can completely solve the harder instances in the underlying benchmarks; however, FSB initially outperforms other policies in achieving a better dual gap for smaller problems; but with time, the slowness factor of FSB kicks in and it falls behind other policies in solving harder instances in terms of dual integral reward. A similar argument applies to the Nuri method in table \ref{tab:rewad_anonymous}. The results suggest that in general our TGAT method generalizes better to the larger instances than other baselines.

\begingroup
\setlength{\tabcolsep}{1.2pt} 
\renewcommand{\arraystretch}{1} 
\begin{table*}[]
\centering
    \caption{Dual Integral Reward for the maritime inventory routing Dataset.}
\vspace{1pt}
\fontsize{5.6}{7.5}\selectfont
{
    \centering
    \begin{tabular}{c}
         \Xhline{2\arrayrulewidth}
        \begin{tabular}{l|cccccccc}
            {Method | Time}& 60s&240s&480s&900s&1200s&2400s&3600s\\    
             \Xhline{1\arrayrulewidth}
             FSB&   \textbf{1,828,117}&	7,084,200&	13,506,208&	24,812,337&	33,215,861&	66,904,807&	100,815,459\\
             \Xhline{1\arrayrulewidth}
             PB&   1,624,580&	6,621,338&	13,392,385&	25,288,001&	33,830,325&	67,957,522&	102,177,927\\
             GCNN\cite{gasse2019exact}&  1,627,863&	6,576,079&	13,248,266&	25,123,252&	33,705,099&	69,139,207&	103,991,280\\
             EL-OROAS\cite{ml4co_banitalebi}&1,682,022&	6,926,643&	14,108,146&	26,743,861&	36,060,519&	73,039,994&	109,997,616\\
             Nuri\cite{ml4co_Nuri}& 1,744,715&	\textbf{7,168,752}&	\textbf{14,516,105}&	\textbf{27,464,789}&	36,760,604&	74,077,350&	111,507,638\\
             \Xhline{1\arrayrulewidth}
             GAT$^{\dagger}$(32,2)&1,690,743&	6,971,475&	14,163,826&	26,606,932&	36,171,609&	73,413,862&	109,926,566\\
             TGAT$^{\dagger}$(32,2,4)& 1,732,502&	7,118,571&	14,414,492&	27,272,535&	\textbf{37,164,971}&	\textbf{74,892,201}&	\textbf{112,934,222}
        \end{tabular}
    \end{tabular}
    \label{tab:rewad_anonymous}
}
\end{table*}
\endgroup

\section{Conclusion}\label{conclusion}
In this paper, we proposed to encode the bipartite graph representation of a MILP instance with two successive passes of the graph attention message passing network. We argued that through the attention mechanism, we can better represent both the feature level and structure level importance of the neighboring nodes. Later, we proposed to encode the temporal correlations of the node embeddings with a GRU structure. We reason that the past states of the graph embeddings contain information that can be used in the current branching episode. By experiments on 6 different datasets that are challenging for state-of-the-art solvers, we corroborate the validity of our proposed method. The experiment results show that in general, our temporo-attentional method generalizes better on larger MILP instances with more complex structures. We hope our work can facilitate further research on incorporating the attention and temporal mechanisms of MILPs into modern combinatorial optimization solvers.

\section{Statement of Ethics}\label{ethics}
This paper does not introduce a new dataset, nor it leverages any personal data.


\FloatBarrier

{\small
\bibliographystyle{splncs04}
\bibliography{references}

\begin{thebibliography}{10}
\providecommand{\url}[1]{\texttt{#1}}
\providecommand{\urlprefix}{URL }
\providecommand{\doi}[1]{https://doi.org/#1}

\bibitem{achterberg2009scip}
Achterberg, T.: Scip: solving constraint integer programs. Mathematical
  Programming Computation  \textbf{1}(1),  1--41 (2009)

\bibitem{achterberg2005branching}
Achterberg, T., Koch, T., Martin, A.: Branching rules revisited. Operations
  Research Letters  \textbf{33}(1),  42--54 (2005)

\bibitem{achterberg2013mixed}
Achterberg, T., Wunderling, R.: Mixed integer programming: Analyzing 12 years
  of progress. In: Facets of combinatorial optimization, pp. 449--481. Springer
  (2013)

\bibitem{alvarez2017machine}
Alvarez, A.M., Louveaux, Q., Wehenkel, L.: A machine learning-based
  approximation of strong branching. INFORMS Journal on Computing
  \textbf{29}(1),  185--195 (2017)

\bibitem{ml4co_banitalebi}
Banitalebi-Dehkordi, A., Zhang, Y.: Ml4co: Is gcnn all you need? graph
  convolutional neural networks produce strong baselines for combinatorial
  optimization problems, if tuned and trained properly, on appropriate data.
  arXiv preprint arXiv:2112.12251  (2021)

\bibitem{bennell2017dynamic}
Bennell, J.A., Mesgarpour, M., Potts, C.N.: Dynamic scheduling of aircraft
  landings. European Journal of Operational Research  \textbf{258}(1),
  315--327 (2017)

\bibitem{bergman2016decision}
Bergman, D., Cire, A.A., Van~Hoeve, W.J., Hooker, J.: Decision diagrams for
  optimization, vol.~1. Springer (2016)

\bibitem{bixby2007gurobi}
Bixby, B.: The gurobi optimizer. Transp. Re-search Part B  \textbf{41}(2),
  159--178 (2007)

\bibitem{boyd2004convex}
Boyd, S., Boyd, S.P., Vandenberghe, L.: Convex optimization. Cambridge
  university press (2004)

\bibitem{breiman2001random}
Breiman, L.: Random forests. Machine learning  \textbf{45}(1),  5--32 (2001)

\bibitem{brody2021attentive}
Brody, S., Alon, U., Yahav, E.: How attentive are graph attention networks?
  arXiv preprint arXiv:2105.14491  (2021)

\bibitem{burges2010ranknet}
Burges, C.J.: From ranknet to lambdarank to lambdamart: An overview. Learning
  \textbf{11}(23-581), ~81 (2010)

\bibitem{ml4co_Nuri}
Cao, Z., Xu, Y., Huang, Z., Zhou, S.: Ml4co-kida: Knowledge inheritance in
  dataset aggregation. arXiv preprint arXiv:2201.10328  (2022)

\bibitem{cornuejols1991comparison}
Cornu{\'e}jols, G., Sridharan, R., Thizy, J.M.: A comparison of heuristics and
  relaxations for the capacitated plant location problem. European journal of
  operational research  \textbf{50}(3),  280--297 (1991)

\bibitem{ding2020accelerating}
Ding, J.Y., Zhang, C., Shen, L., Li, S., Wang, B., Xu, Y., Song, L.:
  Accelerating primal solution findings for mixed integer programs based on
  solution prediction. In: Proceedings of the AAAI Conference on Artificial
  Intelligence. vol.~34, pp. 1452--1459 (2020)

\bibitem{ml4co_competition}
ecole.ai: Ml4co: 2021 neurips competition on machine learning for combinatorial
  optimization. https://www.ecole.ai/2021/ml4co-competition/  (2021)

\bibitem{fiala2013penlab}
Fiala, J., Ko{\v{c}}vara, M., Stingl, M.: Penlab: A matlab solver for nonlinear
  semidefinite optimization. arXiv preprint arXiv:1311.5240  (2013)

\bibitem{flood1956traveling}
Flood, M.M.: The traveling-salesman problem. Operations research
  \textbf{4}(1),  61--75 (1956)

\bibitem{GamrathEtal2020ZR}
Gamrath, G., Anderson, D., Bestuzheva, K., Chen, W.K., Eifler, L., Gasse, M.,
  Gemander, P., Gleixner, A., Gottwald, L., Halbig, K., Hendel, G., Hojny, C.,
  Koch, T., Le~Bodic, P., Maher, S.J., Matter, F., Miltenberger, M.,
  M{\"u}hmer, E., M{\"u}ller, B., Pfetsch, M.E., Schl{\"o}sser, F., Serrano,
  F., Shinano, Y., Tawfik, C., Vigerske, S., Wegscheider, F., Weninger, D.,
  Witzig, J.: {The SCIP Optimization Suite 7.0}. ZIB-Report 20-10, Zuse
  Institute Berlin (March 2020),
  \url{http://nbn-resolving.de/urn:nbn:de:0297-zib-78023}

\bibitem{gamrath2020scip}
Gamrath, G., Anderson, D., Bestuzheva, K., Chen, W.K., Eifler, L., Gasse, M.,
  Gemander, P., Gleixner, A., Gottwald, L., Halbig, K., et~al.: The scip
  optimization suite 7.0  (2020)

\bibitem{ml4co_gasse}
Gasse, M., Cappart, Q., Charfreitag, J., Charlin, L., Chételat, D., Chmiela,
  A., Dumouchelle, J., Gleixner, A., Kazachkov, A.M., Khalil, E., Lichocki, P.,
  Lodi, A., Lubin, M., Maddison, C.J., Morris, C., Papageorgiou, D.J.,
  Parjadis, A., Pokutta, S., Prouvost, A., Scavuzzo, L., Zarpellon, G., Yang,
  L., Lai, S., Wang, A., Luo, X., Zhou, X., Huang, H., Shao, S., Zhu, Y.,
  Zhang, D., Quan, T., Cao, Z., Xu, Y., Huang, Z., Zhou, S., Binbin, C.,
  Minggui, H., Hao, H., Zhiyu, Z., Zhiwu, A., Kun, M.: The machine learning for
  combinatorial optimization competition (ml4co): Results and insights. arXiv
  preprint: 2203.02433  (2022)

\bibitem{gasse2019exact}
Gasse, M., Ch{\'e}telat, D., Ferroni, N., Charlin, L., Lodi, A.: Exact
  combinatorial optimization with graph convolutional neural networks. Advances
  in Neural Information Processing Systems  \textbf{32} (2019)

\bibitem{geurts2006extremely}
Geurts, P., Ernst, D., Wehenkel, L.: Extremely randomized trees. Machine
  learning  \textbf{63}(1),  3--42 (2006)

\bibitem{graham1985history}
Graham, R.L., Hell, P.: On the history of the minimum spanning tree problem.
  Annals of the History of Computing  \textbf{7}(1),  43--57 (1985)

\bibitem{gupta2020hybrid}
Gupta, P., Gasse, M., Khalil, E., Mudigonda, P., Lodi, A., Bengio, Y.: Hybrid
  models for learning to branch. Advances in neural information processing
  systems  \textbf{33},  18087--18097 (2020)

\bibitem{Hamilton_grl}
Hamilton, W.L.: Graph representation learning. Synthesis Lectures on Artificial
  Intelligence and Machine Learning  \textbf{14}(3),  1--159 (2020)

\bibitem{he2014learning}
He, H., Daume~III, H., Eisner, J.M.: Learning to search in branch and bound
  algorithms. Advances in neural information processing systems  \textbf{27}
  (2014)

\bibitem{hoffman2013integer}
Hoffman, K.L., Ralphs, T.K.: Integer and combinatorial optimization.
  Encyclopedia of Operations Research and Management Science pp. 771--783
  (2013)

\bibitem{howard1960dynamic}
Howard, R.A.: Dynamic programming and markov processes. John Wiley (1960)

\bibitem{hussein2017imitation}
Hussein, A., Gaber, M.M., Elyan, E., Jayne, C.: Imitation learning: A survey of
  learning methods. ACM Computing Surveys (CSUR)  \textbf{50}(2),  1--35 (2017)

\bibitem{hutter2011sequential}
Hutter, F., Hoos, H.H., Leyton-Brown, K.: Sequential model-based optimization
  for general algorithm configuration. In: International conference on learning
  and intelligent optimization. pp. 507--523. Springer (2011)

\bibitem{khalil2017learning}
Khalil, E., Dai, H., Zhang, Y., Dilkina, B., Song, L.: Learning combinatorial
  optimization algorithms over graphs. Advances in neural information
  processing systems  \textbf{30} (2017)

\bibitem{khalil2016learning}
Khalil, E., Le~Bodic, P., Song, L., Nemhauser, G., Dilkina, B.: Learning to
  branch in mixed integer programming. In: Proceedings of the AAAI Conference
  on Artificial Intelligence. vol.~30 (2016)

\bibitem{kipf2016semi}
Kipf, T.N., Welling, M.: Semi-supervised classification with graph
  convolutional networks. In: International Conference on Learning
  Representations (ICLR) (2017)

\bibitem{knueven2020mixed}
Knueven, B., Ostrowski, J., Watson, J.P.: On mixed-integer programming
  formulations for the unit commitment problem. INFORMS Journal on Computing
  \textbf{32}(4),  857--876 (2020)

\bibitem{land2010automatic}
Land, A.H., Doig, A.G.: An automatic method for solving discrete programming
  problems. In: 50 Years of Integer Programming 1958-2008, pp. 105--132.
  Springer (2010)

\bibitem{leyton2000towards}
Leyton-Brown, K., Pearson, M., Shoham, Y.: Towards a universal test suite for
  combinatorial auction algorithms. In: Proceedings of the 2nd ACM conference
  on Electronic commerce. pp. 66--76 (2000)

\bibitem{manual1987ibm}
Manual, C.U.: Ibm ilog cplex optimization studio. Version  \textbf{12},
  1987--2018 (1987)

\bibitem{morais2010optimal}
Morais, H., K{\'a}d{\'a}r, P., Faria, P., Vale, Z.A., Khodr, H.: Optimal
  scheduling of a renewable micro-grid in an isolated load area using
  mixed-integer linear programming. Renewable Energy  \textbf{35}(1),  151--156
  (2010)

\bibitem{nair2020solving}
Nair, V., Bartunov, S., Gimeno, F., von Glehn, I., Lichocki, P., Lobov, I.,
  O'Donoghue, B., Sonnerat, N., Tjandraatmadja, C., Wang, P., et~al.: Solving
  mixed integer programs using neural networks. arXiv preprint arXiv:2012.13349
   (2020)

\bibitem{papageorgiou2014mirplib}
Papageorgiou, D.J., Nemhauser, G.L., Sokol, J., Cheon, M.S., Keha, A.B.:
  Mirplib--a library of maritime inventory routing problem instances: Survey,
  core model, and benchmark results. European Journal of Operational Research
  \textbf{235}(2),  350--366 (2014)

\bibitem{pomerleau1991efficient}
Pomerleau, D.A.: Efficient training of artificial neural networks for
  autonomous navigation. Neural computation  \textbf{3}(1),  88--97 (1991)

\bibitem{prouvost2020ecole}
Prouvost, A., Dumouchelle, J., Scavuzzo, L., Gasse, M., Ch{\'e}telat, D., Lodi,
  A.: Ecole: A gym-like library for machine learning in combinatorial
  optimization solvers. In: Learning Meets Combinatorial Algorithms at
  NeurIPS2020 (2020), \url{https://openreview.net/forum?id=IVc9hqgibyB}

\bibitem{ml4co_qu}
Qu, Q., Li, X., Zhou, Y.: Yordle: An efficient imitation learning for branch
  and bound. arXiv preprint arXiv:2112.12251  (2022)

\bibitem{ross2011reduction}
Ross, S., Gordon, G., Bagnell, D.: A reduction of imitation learning and
  structured prediction to no-regret online learning. In: Proceedings of the
  fourteenth international conference on artificial intelligence and
  statistics. pp. 627--635. JMLR Workshop and Conference Proceedings (2011)

\bibitem{salkin1975knapsack}
Salkin, H.M., De~Kluyver, C.A.: The knapsack problem: a survey. Naval Research
  Logistics Quarterly  \textbf{22}(1),  127--144 (1975)

\bibitem{shen2021learning}
Shen, Y., Sun, Y., Eberhard, A., Li, X.: Learning primal heuristics for mixed
  integer programs. In: 2021 International Joint Conference on Neural Networks
  (IJCNN). pp.~1--8. IEEE (2021)

\bibitem{song2018learning}
Song, J., Lanka, R., Zhao, A., Bhatnagar, A., Yue, Y., Ono, M.: Learning to
  search via retrospective imitation. arXiv preprint arXiv:1804.00846  (2018)

\bibitem{vaz2009pswarm}
Vaz, A.I.F., Vicente, L.N.: Pswarm: a hybrid solver for linearly constrained
  global derivative-free optimization. Optimization Methods \& Software
  \textbf{24}(4-5),  669--685 (2009)

\bibitem{velickovic2017graph}
Velickovic, P., Cucurull, G., Casanova, A., Romero, A., Lio, P., Bengio, Y.:
  Graph attention networks. stat  \textbf{1050}, ~20 (2017)

\bibitem{xu2015empirical}
Xu, B., Wang, N., Chen, T., Li, M.: Empirical evaluation of rectified
  activations in convolutional network. arXiv preprint arXiv:1505.00853  (2015)

\bibitem{rl4co}
Zhang, T., Banitalebi-Dehkordi, A., Zhang, Y.: Deep reinforcement learning for
  exact combinatorial optimization: Learning to branch. 26th International
  Conference on Pattern Recognition, ICPR  (2022)

\end{thebibliography}
}


%
%
%
%





\end{document}